\documentclass[letterpaper]{article} 
\usepackage{aaai2026}  
\usepackage{times}  
\usepackage{helvet}  
\usepackage{courier}  
\usepackage[hyphens]{url}  
\usepackage{graphicx} 
\usepackage{natbib}  
\usepackage{caption} 
\usepackage{algorithm}
\usepackage{algorithmic}

\usepackage{newfloat}
\usepackage{listings}
\DeclareCaptionStyle{ruled}{labelfont=normalfont,labelsep=colon,strut=off} 
\floatstyle{ruled}
\newfloat{listing}{tb}{lst}{}
\floatname{listing}{Listing}
\title{A Text-Routed Sparse Mixture-of-Experts Model with Explanation and Temporal Alignment for Multi-Modal Sentiment Analysis}
\author {
	Dongning Rao\textsuperscript{\rm 1},
	Yunbiao Zeng\textsuperscript{\rm 1, \rm 2},
	Zhihua Jiang\textsuperscript{\rm 3}\footnotemark[1],
	Jujian Lv\textsuperscript{\rm 2}\thanks{Corresponding authors who contributed equally.}\\ 
}
\affiliations {
	\textsuperscript{\rm 1} School of Computer, Guangdong University of Technology, Guangzhou 510006, China\\
	\textsuperscript{\rm 2} School of Computer Science, Guangdong Polytechnic Normal University\\
	\textsuperscript{\rm 3} Department of Computer Science, Jinan University, Guangzhou 510632, China\\
	raodn@gdut.edu.cn, 2112405257@mail2.gdut.edu.cn, tjiangzhh@jnu.edu.cn, jujianlv@gpnu.edu.cn
}

\usepackage{bibentry}
\newcommand{\p}{MSA}
\newcommand{\m}{TEXT}
\newcommand{\dsm}{MOSI}
\newcommand{\dse}{MOSEI}
\newcommand{\dsc}{CH-SIMS}
\newcommand{\dscc}{CH-SIMSv2}
\newcommand{\eq}{Eq. }

\newcommand{\fig}{Fig. }

\newcommand{\tab}{Table. }

\usepackage{mathrsfs}
\usepackage{amssymb}
\usepackage{listings,lstautogobble}
\usepackage{pifont}
\usepackage{threeparttable}
\usepackage[table]{xcolor}
\usepackage{multirow}
\usepackage{tcolorbox}
\usepackage{subfigure}
\usepackage{enumitem}
\usepackage{circledsteps}

\lstdefinelanguage{PROMPT}
{morekeywords={Definition, Input, input,  Output, output, example, Example}
}
\lstdefinestyle{styleAngle}{
	language=PROMPT,
	stringstyle=\color{mauve},
	moredelim=[s][\color{violet}]{\[}{\]}, 
	moredelim=[s][\color{brown}]{<}{>} ,
}

\begin{document}

\twocolumn[{%
	\renewcommand\twocolumn[1][]{#1}%
	\maketitle
	
}]

\insert\footins{\noindent\footnotesize \par *Corresponding authors who contributed equally.\\
	Copyright \copyright\space 2026,
	Association for the Advancement of Artificial Intelligence (www.aaai.org).
	All rights reserved.}

\begin{abstract}
Human-interaction-involved applications	underscore the need for \textbf{M}ulti-modal \textbf{S}entiment \textbf{A}nalysis (\p). Although many approaches have been proposed to address the subtle emotions in different modalities, the power of explanations and temporal alignments is still underexplored. Thus, this paper proposes the \textbf{T}ext-routed sparse mixture-of-\textbf{E}xperts model with e\textbf{X}planation and \textbf{T}emporal alignment for \p\ (\m). \m\ first augments explanations for \p\ via \textbf{M}ulti-modal \textbf{L}arge \textbf{L}anguage \textbf{M}odels (MLLM), and then novelly aligns the representations of audio and video through a temporality-oriented neural network block. \m\ aligns different modalities with explanations and facilitates a new text-routed sparse mixture-of-experts with gate fusion. Our temporal alignment block merges the benefits of Mamba and temporal cross-attention. As a result, \m\ achieves the best performance across four datasets among all tested models, including three recently proposed approaches and three MLLMs. TEXT wins on at least four metrics out of all six metrics. For example, \m\ decreases the mean absolute error to 0.353 on the \dsc\ dataset, which signifies a  13.5\% decrement compared with recently proposed approaches. 
\end{abstract}
\begin{links}
	\link{Code}{https://github.com/fip-lab/TEXT}
\end{links}

\section{Introduction}
\label{sec:intro}
Applications in healthcare and human-computer interaction rely heavily on multi-modal sentiment analysis. Thus,  popular datasets for \p~ like \dsm~\cite{zadeh2016multimodal}, are proposed. However, more  comprehensive approaches are needed to understand the subtle emotional nuances conveyed in audio and video~\cite{wu2025enriching}.

\begin{figure*}[t]
	\centering
	\includegraphics[width=.99\textwidth]{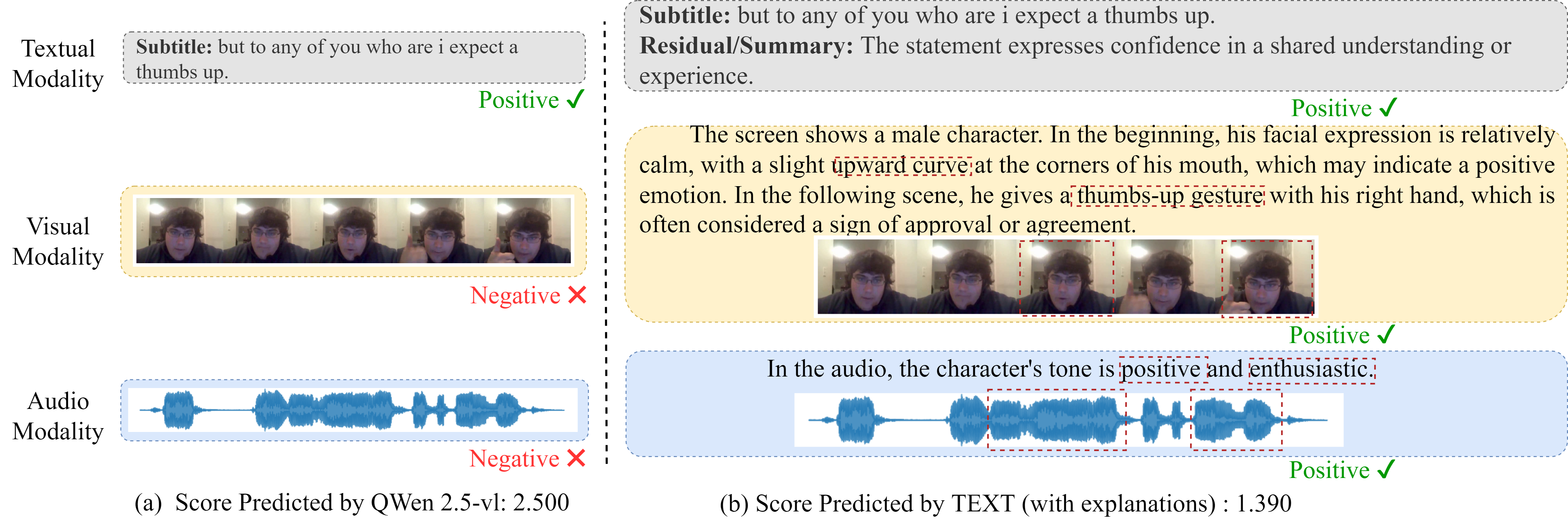}
	\captionof{figure}{An example of \p\ from \dsm. On the left side, from top to bottom, are the textual, visual, and audio modalities of the short video. The corresponding modalities with MLLM-generated explanations are on the right. On the left side, only text can correctly predict the polarity, with a predicted score of 2.500. As a comparison, the label of this example is 1.400, and the \m's prediction is 1.390. With explanations, both video and audio modalities can correctly predict the polarity.}
	\label{fig:demo1}
\end{figure*}

The left part of \fig\ref{fig:demo1} is an example from \dsm, where \p\ demands us to predict not only the polarity but also a score for short videos. For this example, only the text modality can correctly predict the polarity, and the estimated score using all modalities from previous studies~\cite{zhang2023learning} is 1.080. Considering that the label of this sample is 1.400 (deviation: 0.320), there is still room for improvement. 

While fusion is the key to a comprehensive understanding~\cite{zhang2023learning}, recent studies notice that different modalities contribute disparately to \p~\cite{wu2025enriching}. For example, there is always a dominant modality~\cite{feng2024knowledge} (e.g., the text in \fig\ref{fig:demo1}) and text-guide fusion is promising~\cite{wu2025enriching}. However, as two-thirds of the modalities might cause misjudgment in \fig\ref{fig:demo1}, we posit that alignment is the crucial link between representation learning and multimodal fusion. Moreover, despite the aforementioned advances, the power of text in this large language model (LLM)~\cite{bai2025qwen2} era has not been fully explored.

Therefore, we propose the \textbf{T}ext-routed sparse mixture-of-\textbf{E}xperts model with e\textbf{X}planation and \textbf{T}emporal alignment for multi-modal sentiment analysis (\m) in this paper. 
\begin{itemize}
	\item To explore the power of text, \m\ first facilitates multi-modal LLMs (MLLM) to generate explanations that will be encoded by BERT~\cite{devlin2019bert}. The MLLM is VideoLLaMA 3~\cite{zhang2025videollama}, which is fine-tuned with the EMER-fine dataset~\cite{lian2025affectgpt}.
	\item Then, considering how video or audio can mislead (such as in \fig\ref{fig:demo1}), this study aligns audio (Librosa~\cite{mcfee2015librosa}) and video (OpenFace~\cite{baltrusaitis2018morency}) encoding with explanations for the first time. This module uses a \textbf{C}ross-\textbf{A}ttention(CA)-based alignment.
	\item To improve temporal fusion, we propose a novel temporal alignment between the aligned audio/video representations. This block combines, simplifies, and outperforms Mamba~\cite{mamba2} and temporal CA~\cite{zhang2024enhanced}.
	\item Then, for better decisions, we implement a text-routed sparse mixture-of-experts (SMoE)~\cite{shazeer2017outrageously} because of the dominance of text. That is, experts are activated based on the text.
	\item  At last, a multi-layer perceptron (MLP) with gate fusion (GF) ~\cite{qiu2025gated} is applied as a classifier.
\end{itemize}

The right part of \fig\ref{fig:demo1} is the result of \m. With explanations, both visual and audio modalities can correctly predict the polarity. Furthermore, the final score given by \m\ is 1.390, which is close to 1.400 (deviation: 0.010). By contrast, the representative MLLM, QWen 2.5-vl~\cite{bai2025qwen2} predicts a score of 2.500 (deviation: 1.100). The deviation is significantly reduced from 1.100 to 0.010, indicating a substantial improvement in prediction accuracy.

To test the effectiveness of \m, we compare \m\ with three recently proposed models and three MLLMs on four datasets. The three compared models are ALMT~\cite{zhang2023learning}, KuDA~\cite{feng2024knowledge},, and DEVA~\cite{wu2025enriching}. The three MLLMs are Qwen2.5-vl~\cite{bai2025qwen2}, GPT-4o~\cite{achiam2023gpt} and VideoLlama3-7B~\cite{zhang2025videollama}; the four datasets are \dsm~\cite{zadeh2016multimodal}, \dse~\cite{zadeh2018multimodal}, \dsc~\cite{yu2020ch}, and \dscc~\cite{liu2022make}. Experiment results show that \m\ outperforms all compared models. For example, on \dsc\ \m\ improve the mean absolute error (MAE) from 0.449 (ALMT), 0.408 (KuDA), and 0.424 (DEVA) to 0.353 (i.e., a decrease of 13.5\%). Further, our ablation study suggests that temporal alignment is the most crucial component for MAE.

Our contribution can be summarized as follows:

\begin{enumerate}

	\item \m\ aligns encoded audio and video through a novel temporality-oriented neural network block;

	\item \m\ first augments data for \p\ via MLLM with explanations that can be aligned with audio and video;

	\item \m\ uses a new text-routed SMoE;

	\item \m\ is the winner among all tested models on four datasets across six metrics.

\end{enumerate}

\section{Related Work}
\label{sec:rel}
\subsection{Multi-modal Sentiment Analysis}
\label{sec:msa}
MSA has been investigated for its attractive applications like fraud detection~\cite{park2021correspondence}  trading systems~\cite{chen2021sentiment}, human-machine interaction~\cite{rozanska2019multimodal}, and health care applications~\cite{shahmining2020}. Existing methods for \p\ can be classified into two categories: representation learning-centered methods (e.g., ALMT and KuDA) and multi-modal fusion-centered methods~\cite{zhang2023learning} (e.g., DEVA). Both KuDA and DEVA are claimed to be state-of-the-art~\cite{feng2024knowledge, wu2025enriching}.

Formally, inputs of \p\ models include text ($t$), audio ($a$), and visual ($v$). Our goal is to fuse the data from different modalities and predict the sentimental polarity $\hat{y}$ along with a sentiment score between [-1, 1] or [-3, 3]. A score greater than, equal to, or less than zero represents positive, neutral, and negative, respectively. As a regression problem, the basic optimization objective function of \p\ is the MSE loss.  



\subsection{Cross-Modal Temporal Fusion}

\label{sec:cmf}

\begin{figure}[t]
	
	\centering
	
	\includegraphics[width=0.45\textwidth]{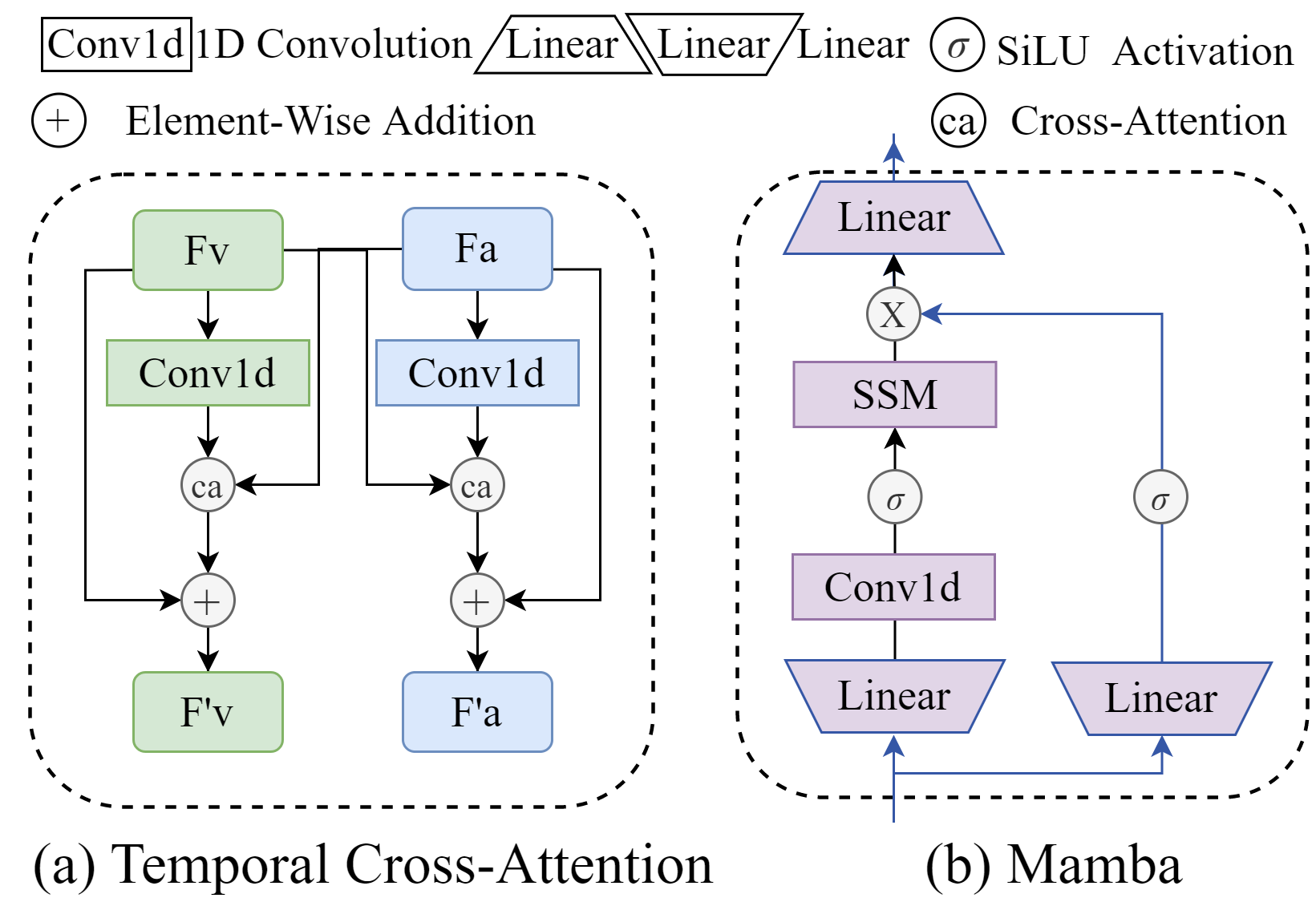} 
	
	\caption{Two temporal alignment block designs: (a) TCA, and (b) Mamba.  Legends are also applicable for \fig\ref{fig:expAlig}$\sim$\ref{fig:tempAlig}. In (a), $F_v$, $F_a$, and $F_t$ represent the input vectors to each module, while $F’_v$, $F’_a$, and $F’_t$ represent the output vectors of the module. In (b), $\otimes$ represents nonlinearity.} 
	
	\label{fig:tca}  
	
\end{figure}



\begin{figure*}[!ht]
	
	\centering
	
	\includegraphics[width=0.98\textwidth]{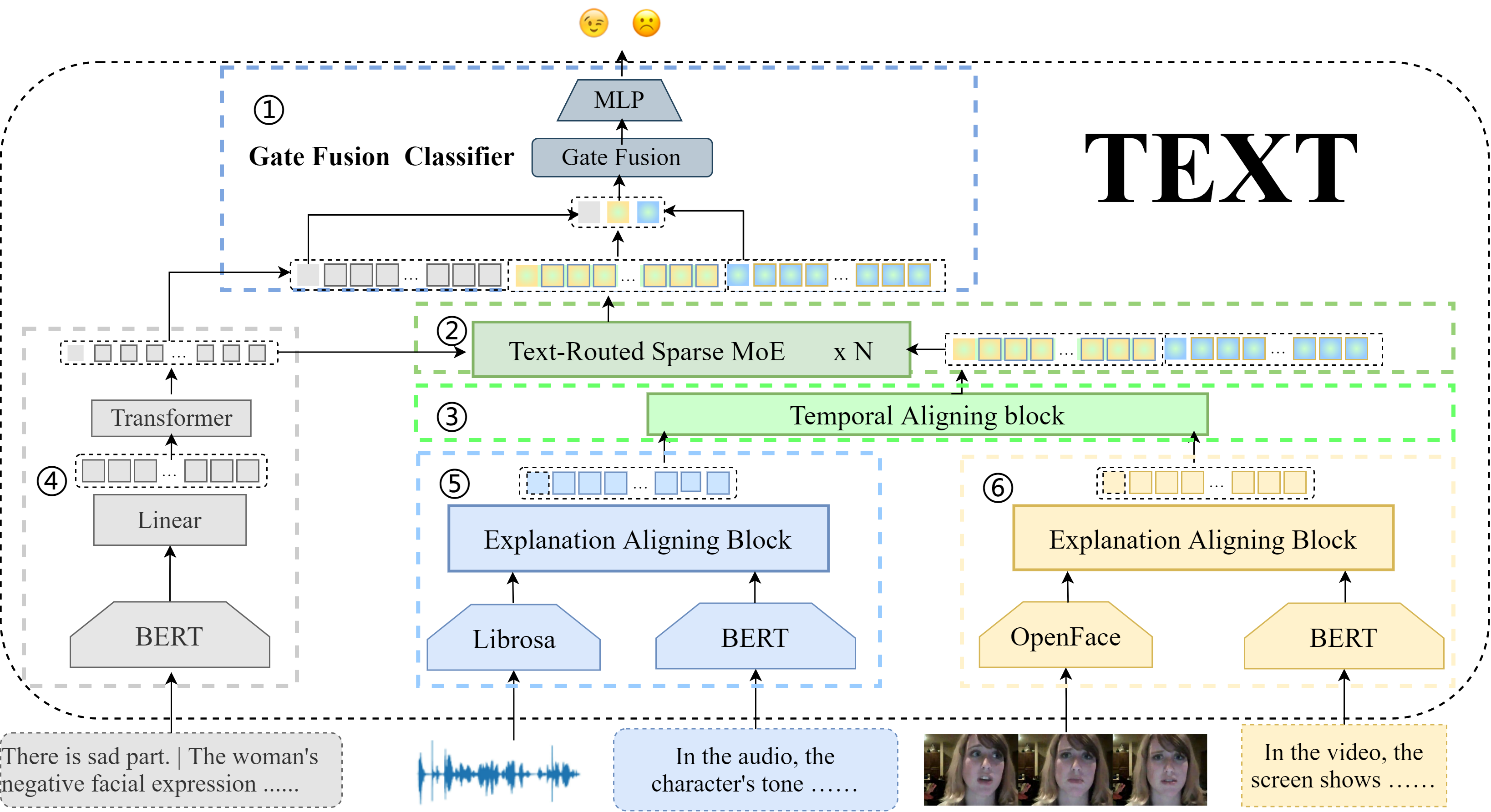} 
	
	\caption{The architecture of \m. From top to bottom, \m\ comprises six functional modules (\ding{172}$\sim$\ding{177}). Its sequential processing workflow centers on three core components: \ding{172} the Gate Fusion Classifier Module, which performs the final decision-making; \ding{173} the Text-Routed SMoE Module, designed to model cross-modal interactions; and \ding{174} the Temporal Alignment Module, responsible for synchronizing audio and video streams. 	In addition, three uni-modal feature extraction modules  (\ding{175}$\sim$\ding{177}) operate in parallel. Notably, both the Audio Feature Extraction Module (\ding{176}) and the Video Feature Extraction Module (\ding{177}) incorporate an Explanation Alignment Block. All modalities are processed using pre-trained encoders: textual data (including explanation annotations) is encoded with BERT, audio signals are extracted using Librosa, and visual information—derived from video frames—is processed via OpenFace.}
	
	\label{fig:arch}
	
\end{figure*}

Because audio and video are sequential, designing neural network blocks for temporal alignment is important. Cross-modal fusion, especially cross-modal attention mechanisms, is a popular temporal alignment approach. Cross-modal fusion captures both similar and dissimilar information between uni-modal representations ($U_m, m \in \{t, v, a \}$) and generates a multi-modal embedding. 

For example, as shown in \fig\ref{fig:tca} (a), the temporal CA (TCA) is a specific emotion-oriented and CA-based block for short video~\cite{zhang2024enhanced}. By contrast, \fig\ref{fig:tca} (b) illustrates the Mamba, which is a linear-time sequence model  with sophisticated \textbf{S}tructured \textbf{S}tate-space \textbf{M}odels (SSM) for long video~\cite{mamba}. However, samples of MSA are often short videos that might exhibit dynamic emotional transitions across frames (e.g., \fig\ref{fig:demo1}). That is, neither Mamba nor TCA is designed for \p. 


\subsection{Sparse Mixture-of-Experts and Gate Fusion}

\label{sec:moe}

To reduce computation overhead and combine multimodal features, two prominent techniques should be considered. First, SMoE~\cite{touvron2023llama} is a technique that avoids unnecessary computation by selectively activating relevant experts. Considering the dominance of text~\cite{wu2025enriching}, we can use text for SMoE routing. Second, GF is a commonly used structure for integrating features from different modalities~\cite{ren2018gated}. That is, we can employ GF to combine text, audio, and video features in \p~\cite{cheng2025one}. However, to our knowledge, neither SMoE nor GF has attracted adequate attention regarding \p.

\section{Method}
\label{sec:model}

\subsection{The Overall Architecture of \m}
\label{sec:arch}
\fig\ref{fig:arch} is the overall architecture of \m. From the top down, \m\ has six modules: \ding{172} GF classifier; \ding{173} text-routed SMoE; \ding{174} temporal aligning; \ding{175} text encoding; \ding{176}\ding{177}: explanation aligning blocks for audio and video (with uni-modal encoding). In this section, we will first introduce our explanation generation approach ($\S$\ref{sec:dataAug}) and then present the uni-modal encoding methods ($\S$\ref{sec:uniEnc}). The explanation for aligning blocks in modules \ding{176} and \ding{177} is provided in $\S$\ref{sec:explAlign}, and details of the temporal aligning blocks can be found in $\S$\ref{sec:tempAlign}. At last, the \ding{172} GF classifier will be stated in $\S$\ref{sec:mlp} after $\S$\ref{sec:smoe}, which presents the \ding{173} text-routed SMoE.

\subsection{Explanation Generation}

\label{sec:dataAug}

\begin{figure}[!ht]
	\centering
	\includegraphics[width=0.45\textwidth]{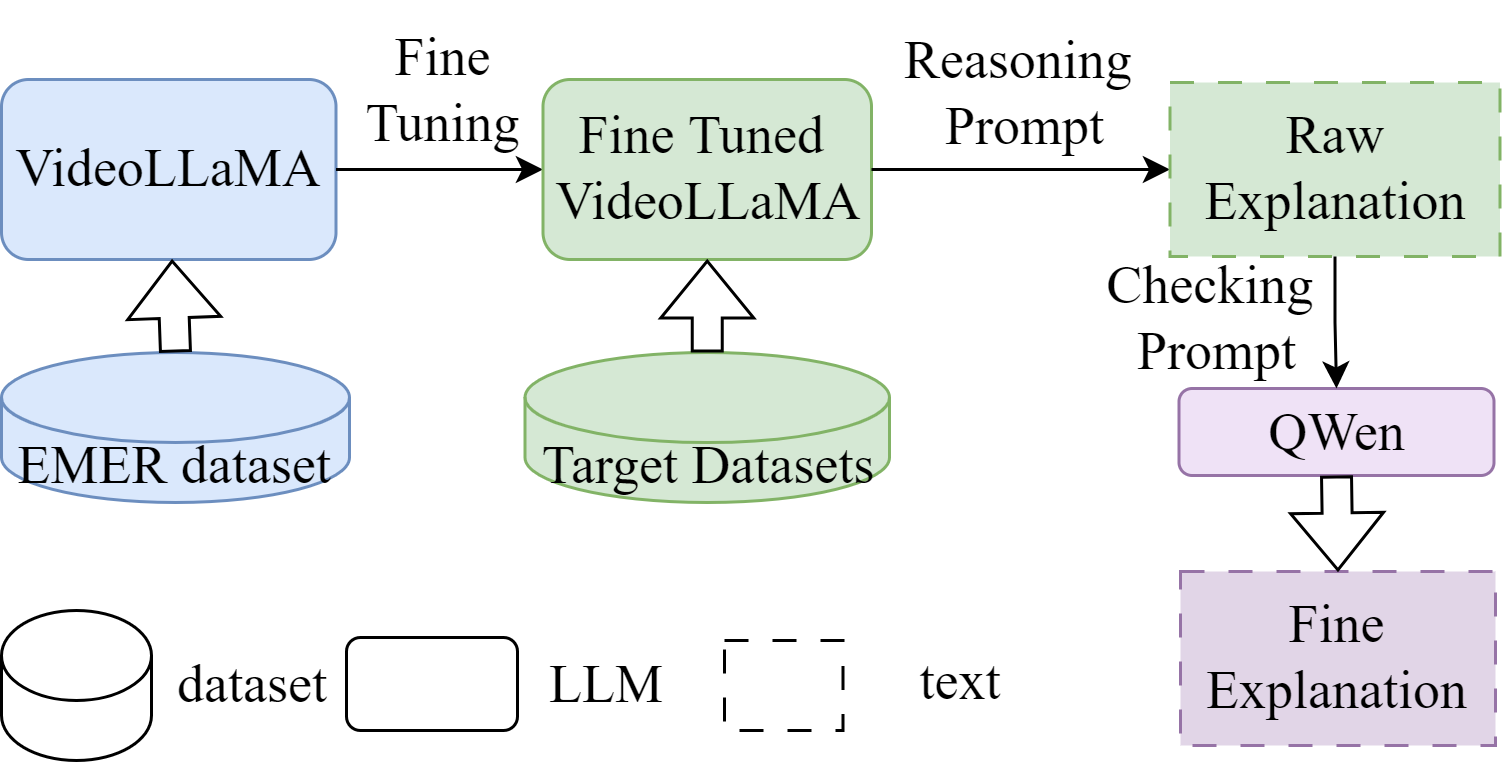} 
	\caption{The procedure of explanation generation.}
	\label{fig:explGenProc}
\end{figure}

To explore the power of MLLM, our explanation generation process is two-stage. See \fig\ref{fig:explGenProc}. We first generate raw explanations using VideoLLaMA 3, which is fine-tuned on the EMER-fine dataset~\cite{lian2025affectgpt}. At this stage, the prompt separates explanations of audio, video, and comments. Then, with Qwen 3, we refine raw explanations to fine explanations with the checking prompt. The prompt for the second stage is called the reasoning prompt.

\subsection{Uni-Modal Encoding}
\label{sec:uniEnc}
We encode subtitles and explanations using BERT, audio using Librosa, and video using OpenFace. See the lower part of \fig\ref{fig:arch}. Viewing models as functions, we have $BERT(t), Librosa(a), OpenFace(v)$ ($B(t), Li(a), OF(v)$ for short). Because we separate the explanation for audio, video, and comments in $\S$\ref{sec:dataAug}, $B(t)$ can be further specified as the explanation for audio $B(e_a)$, the explanation for video $B(e_v)$ and the comments $B(c)$. 

\subsection{Explanation Aligning}
\label{sec:explAlign}

\begin{figure}[t]
	\centering
	\includegraphics[width=0.49\textwidth]{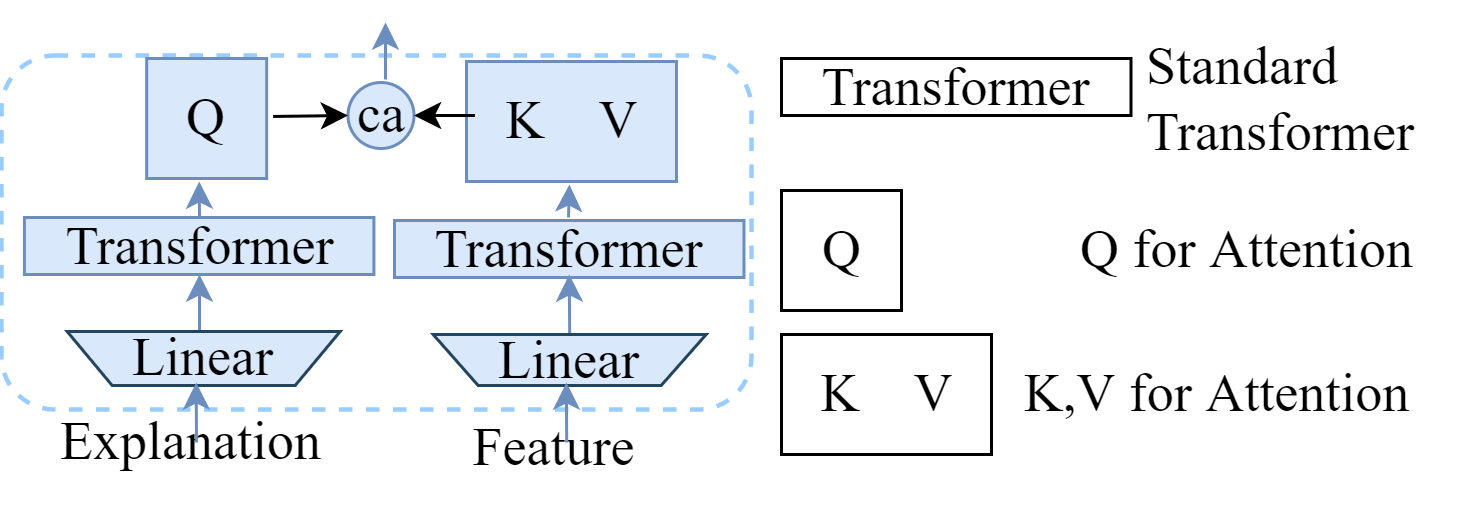} 
	\caption{Explanation aligning block.}
	\label{fig:expAlig}
\end{figure}

In the explanation aligning block, audio $Li(a)$, video $OF(v)$ and subtitles $B(s)$ are aligned with corresponding explanations or comments $B(e_a)$/$B(e_v)$/$B(c)$ via CA ($\Circled{{ca}}(\cdot)$). That is, for feature $F$ and explanation $E$, $\Circled{{ca}}(F, E)$ is as \eq\ref{eq:ct}, where $W_Q, W_K$ and $W_V$ are weights and the transpose of a matrix is $^T$. We illustrate this module in \fig\ref{fig:expAlig}.

\begin{equation}
	\Circled{{ca}}(F, E)= softmax((W_QE)(W_KF)^T)W_VF
	\label{eq:ct}
\end{equation} 

As the lengths of $B(t), Li(a), OF(v)$ may differ, we use 50 tokens for all uni-modal encodes and a learnable token for feature aggregation. The final representation is a 51-dimensional embedding. The results are aligned representations of text, audio, and video. Namely, $E_t, E_a$, and $E_v$.

\subsection{Temporal Aligning}
\label{sec:tempAlign}
\begin{figure}[t]
	\centering
	\includegraphics[width=0.45\textwidth]{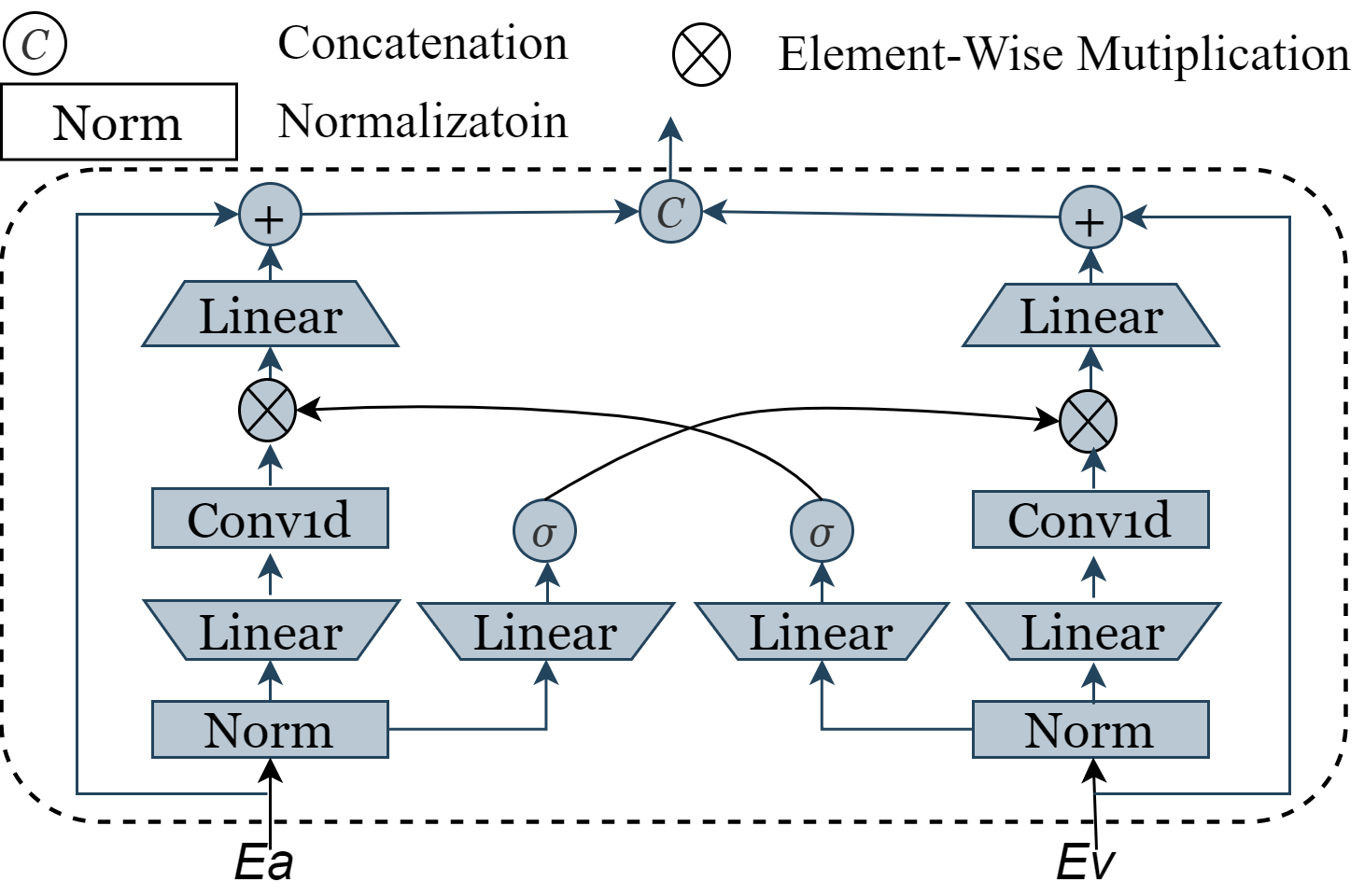} 
	\caption{Temporal aligning block. }
	\label{fig:tempAlig}
\end{figure}

\m\ uses a novel temporal alignment block, which includes one convolution ($Conv1d(\cdot)$) and two linear operations. It combines the advantages of Mamba and TCA. \fig\ref{fig:tempAlig} shows this temporal aligning block. Comparing \fig\ref{fig:tempAlig} with \fig\ref{fig:tca}, we can see that our temporal aligning block is simpler than Mamba and TCA because it does not involve CA or SSM.

Formally, $\otimes$ is used for element-wise multiplication, and $\oplus$ is for element-wise addition. We use $\Circled{c}$ to represent the concatenation of features, let $L(\cdot)$ denote the linear layer, and let $N(\cdot)$ represent the normalization layer. Then, $LN(\cdot)$ is a linear layer after a normalization layer. The Sigmoid Linear Unit activation function (SiLU), $\Circled{\sigma}$, is used for activation. Let \eq\ref{eq:tal} be the left part of \fig\ref{fig:tempAlig}, \eq\ref{eq:tar} be the right part of \fig\ref{fig:tempAlig}, and $E_{av}$ be the temporal aligned representation. The temporal alignment can be defined as  \eq\ref{eq:ta}.

\begin{equation}
	left= E_a \oplus L(Conv1d(LN(E_a)) \otimes  \Circled{\sigma}(LN(E_v))  )
	\label{eq:tal}
\end{equation}

\begin{equation}
	right= E_v \oplus L(Conv1d(LN(E_v)) \otimes \Circled{\sigma}(LN(E_a))  )
	\label{eq:tar}
\end{equation}

\begin{equation}
	E_{av}=\Circled{c}(left, right)
	\label{eq:ta}
\end{equation}

\subsection{Text-Routed SMoE}

\label{sec:smoe}

\begin{figure}[t]
	
	\centering
	
	\includegraphics[width=0.45\textwidth]{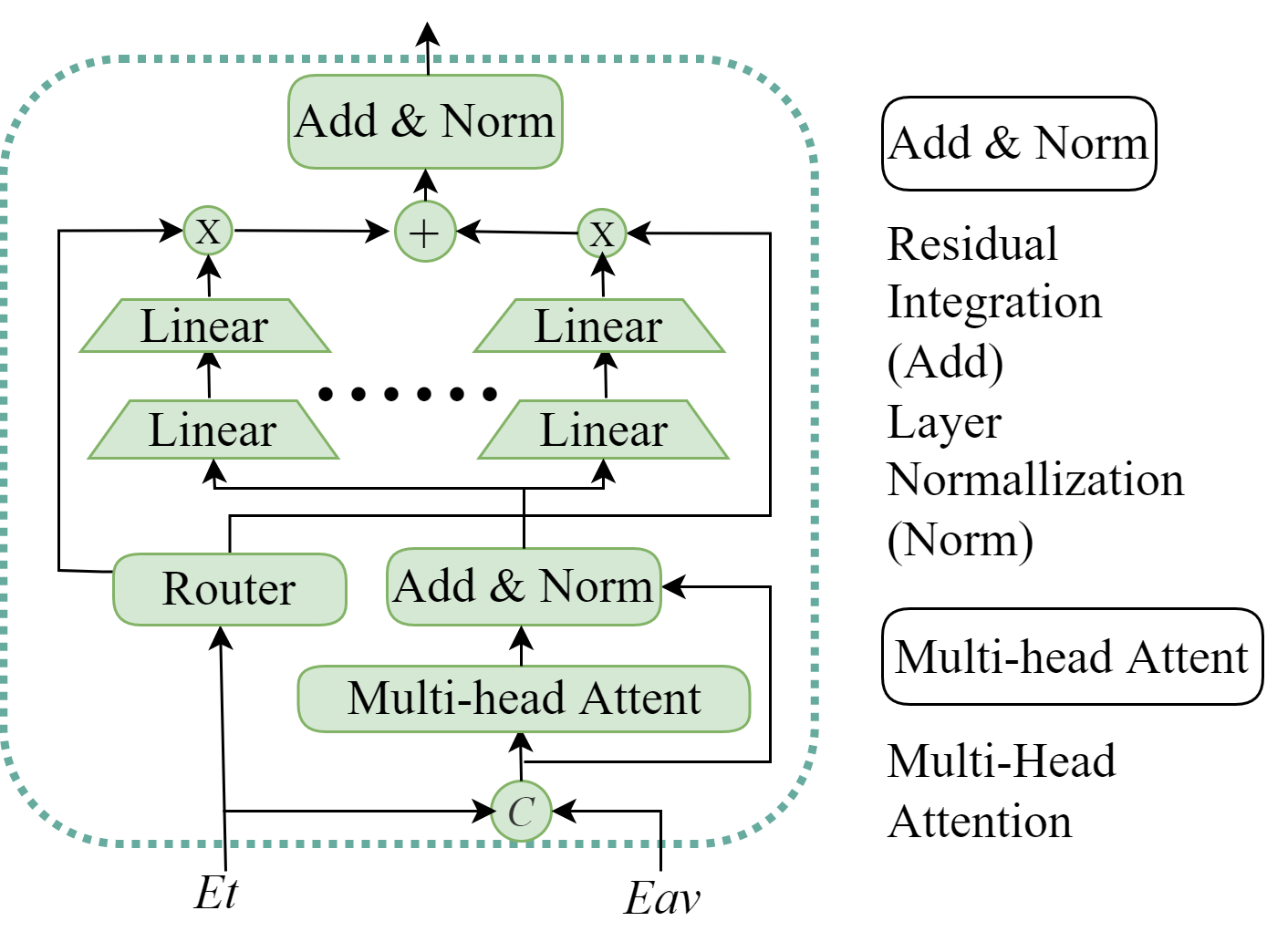} 
	
	\caption{The SMoE block. The symbol $\Circled{x}$ represents multiplication, and “Router” refers to the routing function.}
	
	\label{fig:MoE}
	
\end{figure}

This module is illustrated in \fig\ref{fig:MoE}.

	\begin{table*}\small
		\centering
		\begin{threeparttable}
				\begin{tabular}{
						>{\centering\arraybackslash}p{1.2cm}|
						>{\centering\arraybackslash}p{1.2cm}
						>{\centering\arraybackslash}p{0.8cm}
						>{\centering\arraybackslash}p{0.8cm}
						>{\centering\arraybackslash}p{1.3cm}
						>{\centering\arraybackslash}p{0.6cm}
						>{\centering\arraybackslash}p{0.6cm}|
						>{\centering\arraybackslash}p{1.2cm}
>{\centering\arraybackslash}p{0.8cm}
>{\centering\arraybackslash}p{0.8cm}
>{\centering\arraybackslash}p{1.3cm}
>{\centering\arraybackslash}p{0.6cm}
>{\centering\arraybackslash}p{0.6cm}
					}
					\hline
					\multirow{2}{*}{Model} & \multicolumn{6}{c|}{MOSI} & \multicolumn{6}{c}{MOSEI} \\
					\cline{2-7} \cline{8-13}
								& Acc-2 & Acc-5 & Acc-7 & F1 & MAE$\downarrow$ & Corr & Acc-2 & Acc-5 & Acc-7 & F1 & MAE$\downarrow$ & Corr \\
					\hline
					
					ALMT 		& 83.10/85.23 & 50.41 & 45.01 & 83.20/85.37 & 0.716 & 0.773 & 82.39/85.87 & 53.96 & 52.16 & 82.18/85.95 & 0.542 & 0.767 \\
					KuDA\tnote{1} 		& 84.40/86.43 & N/A & \textbf{47.08} & 84.48/86.46 & 0.705 & 0.795 & 83.26/\underline{86.46} & N/A  & \textbf{52.89} & 82.97/\underline{86.59} & \underline{0.529} & \underline{0.776} \\
					DEVA 		& 84.40/86.29 & 51.78 & \underline{46.32} & 84.48/86.30 & 0.730 & 0.787 & 83.26/86.13 & \textbf{55.32} & 52.26 & 82.93/86.21 & 0.541 & 0.769 \\
					\hline
					
					GPT-4o 		& \underline{85.71}/\underline{86.74} & \underline{52.59} & 44.61 & \underline{85.68}/\underline{86.68} & \underline{0.682} & \underline{0.823} & \underline{84.77}/86.08 & 50.53 & 48.38 & \underline{84.82}/86.08 & 0.637 & 0.744 \\
					Qwen & 83.09/83.38 & 45.63 & 36.30 & 83.09/83.31 & 1.129 & 0.677 & 84.14/84.59 & 41.73 & 40.67 & 84.17/84.64 & 1.007 & 0.587 \\
					VL3 & 67.64/68.45 & 28.72 & 23.76 & 68.30/68.48 & 1.437 & 0.442 & 71.07/71.20 & 33.12 & 31.87 & 71.35/71.59 & 1.141 & 0.349 \\
					\hline
					
					\textbf{\m} & \textbf{86.44}/\textbf{88.72} & \textbf{52.62} & 45.92 & \textbf{86.55}/\textbf{88.76} & \textbf{0.666} & \textbf{0.829} & \textbf{85.02}/\textbf{86.57} & \underline{54.05} & \underline{52.29} & \textbf{85.01}/\textbf{86.85} & \textbf{0.528} & \textbf{0.786} \\
					\hline
				\end{tabular}

						\begin{tablenotes}
				\small
				\item[1] We use the results in previous studies for KuDA~\cite{feng2024knowledge} and DEVA ~\cite{wu2025enriching} in this paper. QWen: Qwen2.5-vl. VL3: VideoLLaMA3.

			\end{tablenotes} 
		\end{threeparttable}
		\caption{Comparison on MOSI and MOSEI Datasets. Acc and F1 are shown in percentage scale. All results in our paper is statistical significant using T-test, i.e. $p< 0.05$. The best results are in bold, and the second best results are underlined. Acc-2 and F1-Score are computed in two settings: negative/non-negative (including zero) and negative/positive (excluding zero).}
		\label{table:comparisonMOSI}
	\end{table*}

An SMoE structure using text as the key for route decisions. Suppose the first parameter of the function $SMoE(\cdot)$ is the key of routing; this layer can be formalized as $SMoE(E_t, E_{av})$. 


\subsection{Gate Fusion Classifier}

\label{sec:mlp}

An MLP with a GF($\Circled{\sigma}$) comprises the classifier of \m, see \eq\ref{eq:final}. See the upper part of \fig\ref{fig:arch} for an intuitive understanding. Specifically, only tokens for feature aggregation (see $\S$\ref{sec:explAlign}) are used for the final decision.

\begin{equation}
	L(\Circled{\sigma}(SMoE(E_t, E_{av})))
	\label{eq:final}
\end{equation}

\section{Experiment}

\label{sec:exp}

\subsection{Experiment Settings}

\label{sec:settings}


\subsection{Datasets}

\label{sec:dataset}

\dsm~\cite{zadeh2016multimodal}, \dse~\cite{zadeh2018multimodal}, \dsc~\cite{yu2020ch} and \dscc~\cite{liu2022make} are four popular datasets of \m. The Multi-modal Opinion-Level Sentiment Intensity (\dsm)  is a collection of YouTube monologues. It contains 2,199 subjective words and video clips, which are artificially labeled as consecutive opinion scores. The CMU Multi-modal Opinion Sentiment and Emotion Intensity (\dse) is an improvement on \dsm. It contains 22,856 YouTube monologues and video segments covering 250 distinct topics from 1,000 distinct speakers. The CHinese SIngle- and Multi-modal Sentiment analysis (\dsc) is a Chinese \m\ dataset with fine-grained annotations of modality. It contains 2,281 human-labeled video clips collected from various sources, along with a sentiment score ranging from -1 (strongly negative) to 1 (strongly positive). At last, the \dscc\ is the updated version of \dsc.

\subsection{Compared Models}

\label{sec:comparedModel}

\m\ is compared with three models and three MLLMs. 

The three recently proposed representative models are ALMT~\cite{zhang2023learning}, KuDA~\cite{feng2024knowledge} and DEVA~\cite{wu2025enriching}. First, ALMT learns an irrelevance/conflict-suppressing representation from visual and audio features, and each modality is first transformed into a unified form by using a Transformer~\cite{vaswani2017attention} with initialized tokens. Second, KuDA argues that there is always a dominant modality, which is enhanced by sentiment knowledge. Third, DEVA incorporates the text-guided progressive fusion along with an emotional description generator. Many previous studies in this research line that have been compared with ALMT, KuDA, and DEVA is not listed in this paper. 

The three MLLMs are Qwen2.5-vl~\cite{bai2025qwen2}, GPT-4o~\cite{achiam2023gpt} and VideoLlama3-7B~\cite{zhang2025videollama}. 

\subsection{Metrics}

\label{sec:metrics}

Following the previous works~\cite{zhang2023learning,feng2024knowledge,wu2025enriching}, we facilitate metrics from two classes. The first class is for classification, which includes the Weighted F1 score (F1-Score), binary classification accuracy (Acc-2), three-class classification accuracy (Acc-3), five-class classification accuracy (Acc-5), and seven-class classification accuracy (Acc-7). The second class focuses on regression, which includes the Mean Absolute Error (MAE) and Pearson correlation (r). For all metrics, except MAE, higher values indicate better performance. 

For \dsm\ and \dse, we further compute Acc-2 and F1-Score in two settings, as in previous works~\cite{zhang2023learning,feng2024knowledge,wu2025enriching}. That is, negative/non-negative (including zero) and negative/positive (excluding zero). Further, we calculate Acc-3 and Acc-5 on \dsc\ and \dscc.

\begin{figure}[t]
	
	\centering
	
	\includegraphics[width=0.49\textwidth]{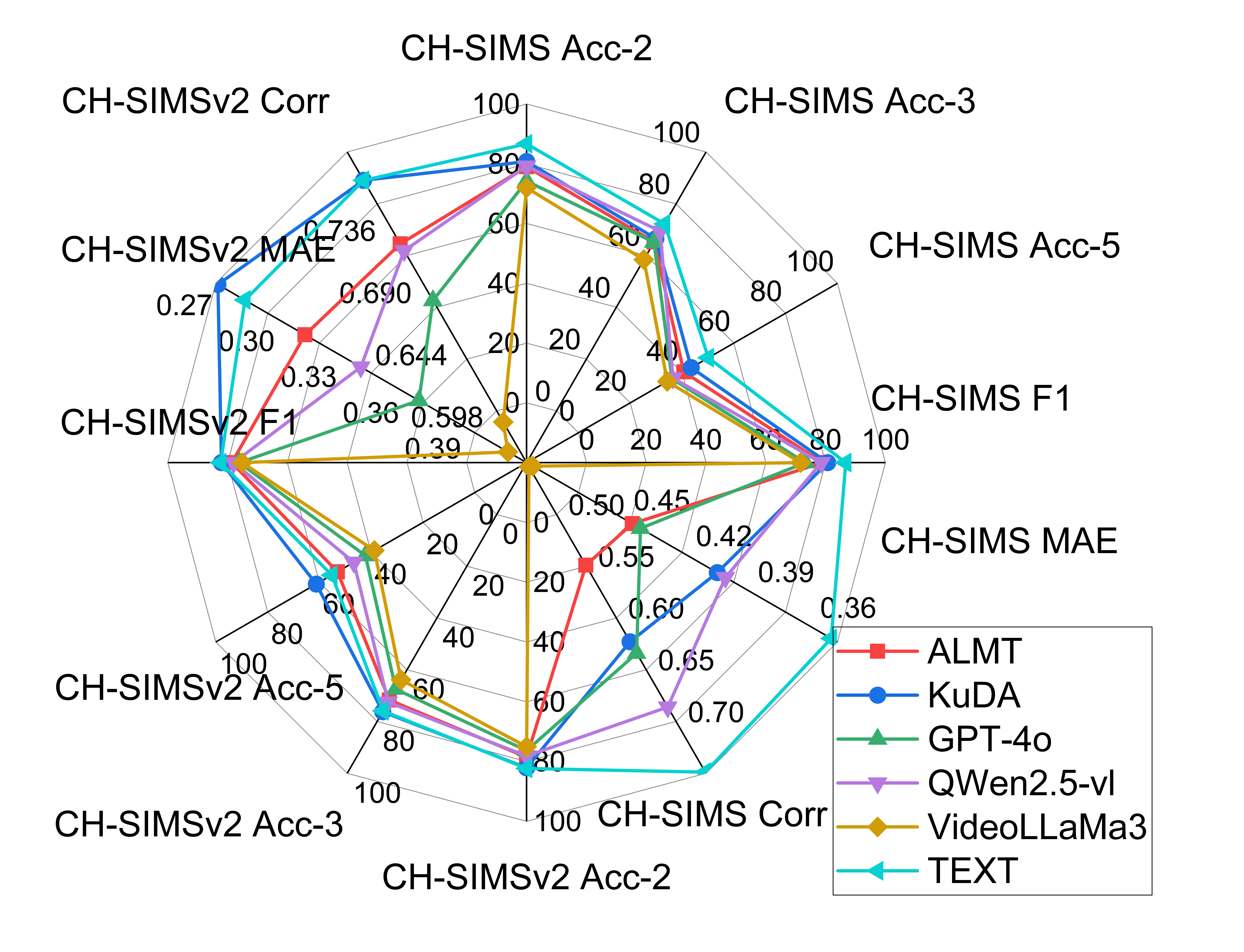} 
	
	\caption{Comparison on CH-SIMS and CH-SIMSv2.  Axes are metrics, and lines are compared models. 
	}
	
	\label{fig:radar}
	
\end{figure}


	\begin{table*}\small
		\centering
		\begin{threeparttable}
			\begin{tabular}{p{2.5cm}|p{2.8cm}|>{\centering}p{1.6cm}>{\centering}p{0.95cm}>{\centering}p{0.95cm}>{\centering}p{1.6cm}>{\centering}p{0.80cm}>{\centering\arraybackslash}p{0.80cm}}
				\hline
				Settings & Method  & Acc-2 & Acc-5 & Acc-7 & F1 & MAE & Corr \\
				\hline
				
				\multirow{8}{*}{w explanations} & \textbf{TEXT}                      & \textbf{85.02}/\textbf{86.57} & 54.05 & 52.29 & \textbf{85.01}/\textbf{86.85} & \textbf{0.528} & \textbf{0.786} \\
				& \qquad A \& V               & 75.51/77.71 & 43.76 & 43.57 & 75.02/77.94 & 0.694 & 0.582 \\
				
				& \qquad  T \& A                        & 83.38/85.06 & \underline{54.07} & \underline{52.46} & 83.54/85.57 & 0.542 & 0.773 \\
				& \qquad  T \& V                        & 83.01/85.75 & 50.55 & 48.62 & 83.89/85.94 & 0.566 & 0.776 \\
				& \qquad\qquad T                         & 83.49/86.43 & \textbf{54.56} & \textbf{52.84} & 83.24/86.57 & \underline{0.535} & 0.771 \\
& \qquad\qquad A                         & 75.90/74.00 & 39.88 & 39.30 & 77.15/75.89 & 0.776 & 0.492 \\
& \qquad\qquad V                         & 72.55/68.49 & 39.47 & 39.47 & 76.05/73.11 & 0.807 & 0.363 \\
				
				\hline				
				
				\multirow{8}{*}{w/o explanations}				& TEXT                      & 83.60/86.02 & 52.39 & 50.35 & 83.42/86.20 & 0.569 & 0.776 \\
					&\qquad A \& V              & 70.59/62.71 & 41.36 & 41.36 & 82.21/76.73 & 0.834 & 0.146 \\
	                & \qquad T \& A                        & 83.92/85.50 & 53.44 & 51.73 & 83.82/85.71 & 0.546 & 0.766 \\
					& \qquad T \& V                        & 83.32/85.83 & 50.96 & 48.98 & 83.05/85.94 & 0.573 & 0.765 \\
				& \qquad\qquad T                         & 83.58/85.58 & 52.39 & 50.80 & 83.42/85.77 & 0.549 & 0.763 \\
				&  \qquad\qquad A                         & 71.02/62.85 & 41.27 & 41.27 & 73.06/70.19 & 0.838 & 0.153 \\
				&  \qquad\qquad V                         & 70.94/62.91 & 41.36 & 41.36 & 73.81/70.12 & 0.828 & 0.172 \\

				\hline
				
					\multirow{6}{*}{\shortstack[l]{Component\\ Ablation\tnote{1}}}
				&  EA$\leftarrow$ Linear              		& 84.25/\underline{86.57} & 49.58 & 48.21 & 84.11/86.77 & 0.577 & 0.762 \\  

				&  TA $\leftarrow \Circled{C}$                    & 83.77/85.42 & 49.52 & 48.40 & 83.83/85.80 & 0.580 & 0.749 \\
 				&  TA $\leftarrow $   Mamba		& \underline{84.80}/86.41 & 52.65 & 50.65 & \underline{84.75}/\underline{86.63} & 0.562 & 0.780  \\
 				&  TA $\leftarrow $ TCA 	&  83.41/86.43 & 53.98 & 51.38 & 83.12/86.55 & 0.565 & \underline{0.781} \\

 &  SMoE $\leftarrow$ Trans  & 83.73/85.33 & 52.22 & 50.29 & 83.70/85.62 & 0.573 & 0.769 \\

				& TEXT   w/o Gating                   & 84.40/86.35 & 51.41 & 49.07 & 84.35/86.62 & 0.571 & 0.780 \\
				\hline

			\end{tabular}
			\begin{tablenotes}
				\small
				\item[1] EA $\leftarrow$ Linear : EA is replaced by a linear layer. TA $\leftarrow \Circled{c}$: TA is replaced by concatenation.  TA $\leftarrow$ Mamba: TA is replaced by Mamba. TA $\leftarrow$ TCA: TA is replaced by TCA. SMoE $\leftarrow$ Trans : SMoE is replaced by the Transformer.				
			\end{tablenotes} 
		\end{threeparttable}
		\caption{Ablation study results on MOSEI. Acc and F1 are shown in percentage scale, and best results are in bold. Acc-2 and F1-Score are computed in two settings: negative/non-negative (including zero) and negative/positive (excluding zero). A: audio; V: video; T: text.  TA: Temporal Alignment; EA: Explanation Alignment.  SMoE: Text-Routed Sparse Mixture-of-Experts.}
		\label{table:ablationMOSEI}
	\end{table*}

\subsection{Model Comparison}

\label{sec:expComparing}

\tab\ref{table:comparisonMOSI} compared our model with compared methods on \dsm\ and \dse. The same comparisons on \dsc\ and \dscc\ are illustrated in \fig\ref{fig:radar}.

\m’s advantages are clear in \tab\ref{table:comparisonMOSI} and \fig\ref{fig:radar}, as it excels in almost every test. Specifically, five key results highlight the significance of \m:

\begin{enumerate}
	
	\item  For Acc-5,  \m\ is better than KuDA (the $2^{nd}$ best model) on \dsc, with 50.15\% accuracy versus KuDA’s 43.54\%.
	
	\item For the F1 score on \dsc, \m\ reaches 86.75\%, maintaining a clear lead over KuDA’s 80.71\%.

	\item Regarding MAE on \dsc, \m\ attains a notably lower error rate of 0.353—indicating superior performance—compared to QWen’s 0.404.

	\item  In the Acc-2 evaluation on the \dsm\ dataset, \m\ achieves 88.72\%, surpassing GPT-4o’s 86.74\%.

	\item For the F1 score on \dsm, \m\ further solidifies its advantage with a score of 88.76\%, outperforming GPT-4o’s 86.68\%.
	
\end{enumerate}

However, there are two exceptions. First, KuDA is the best model for Acc-7 on \dsm\ and \dse. KuDA’s success has also been proven for Acc-5 on \dscc. Second, DEVA shows its advantages on Acc-5 on \dse.  We conjecture that fine-grained scoring requires additional knowledge of sentiment (even domain-specific) to be effective.

\subsection{Ablation Study}

\label{sec:expAblation}

\begin{figure}[t]
	
	\centering
	
	\includegraphics[width=0.49\textwidth]{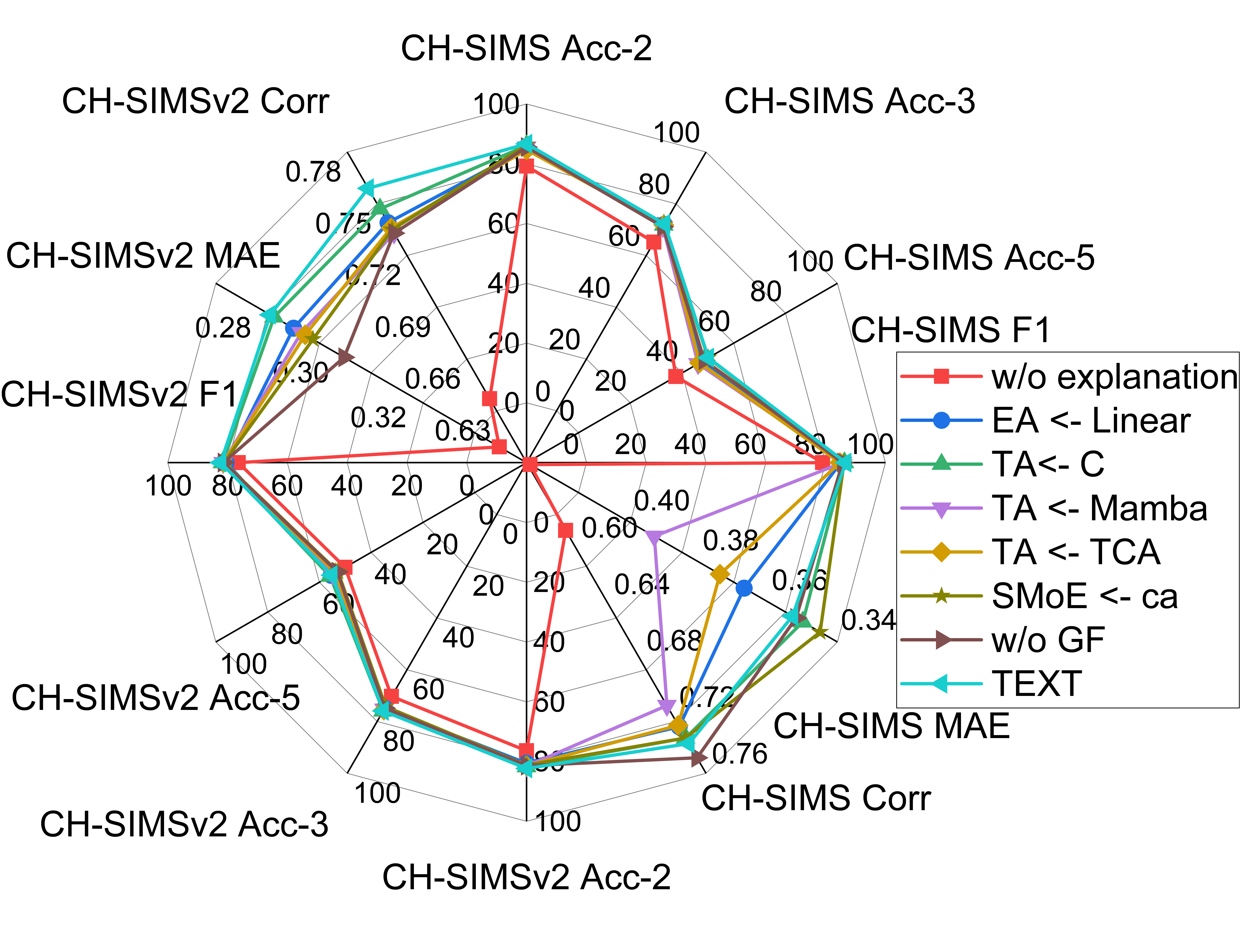} 
	
	\caption{Ablation study on CH-SIMS and CH-SIMSv2. Axes are metrics, and lines are compared models. 
	}
	
	\label{fig:radarAbl}
	
\end{figure}

\tab\ref{table:ablationMOSEI} lists the results of the ablation study on \dse\ and \fig\ref{fig:radarAbl} shows the corresponding results on \dsc\ and \dscc. 

Analysis of \tab\ref{table:ablationMOSEI} yields five key insights into multimodal performance and the effects of ablation studies:

\begin{enumerate}
	
	\item Text as the dominant uni-modal input: it outperforms audio and video across metrics like Acc-5 and Acc-7. For example, regarding Acc-7 on \dse\, using text with explanations yields 52.84\%, while \m\ achieves only 52.29\%. The dominance indicates that the text likely contains more accurate information than other modalities (see \fig\ref{fig:demo1}).
	
	\item Audio and video contribute equally to performance: when integrated with text, both audio and video provide comparable performance boosts. That is, the results of “\verb|T & V|” and “\verb|T & A|” in \tab\ref{table:ablationMOSEI} are similar. These modalities complement textual information with distinct yet equally valuable contextual cues, such as prosody in audio and visual signals in video.
	
	\item Explanations enhance prediction: the removal of explanations results in an approximate 2\% performance decline across most metrics. This drop corroborates earlier findings, suggesting that explanations play a crucial role in integrating text with audio and video.
	
	\item Temporal alignment is crucial for multimodal integration: replacing our temporal alignment block with Mamba or TCA also causes a roughly 2\% performance drop. For example, when we replaced temporal alignment with concatenation, the MAE was 0.580 in \tab\ref{table:ablationMOSEI}. This suggests temporal alignment is crucial for MAE. That is, matching audio and text timing might be key for \p.
	
	\item SMoE demonstrates comparable value to explanations. We believe the effectiveness of SMoE stems from its keyword-sensitive expert activation mechanism. That is, some experts are trained for a specific topic, and text including corresponding keywords will activate these experts. Furthermore, it may foster cross-modal consistency for improved interpretability.

\end{enumerate}



\subsection{Qualitative Examples}

\label{sec:expCase}

\tab\ref{table:sample}  shows the predicted scores evaluated by different settings of \m\ for the case in \fig\ref{fig:demo1}. The third column represents the predicted score, and the last column shows the deviation between the label and the evaluated score.

	\begin{table}\small
		\centering
		\begin{threeparttable}
			\begin{tabular}{p{1.5cm}|p{2.5cm}|p{0.9cm}|p{0.9cm}}
				\hline
				Settings & Model & Score & $\sigma$ $\downarrow$ \\
				\hline
				Human	&	N/A  & 1.400 & N/A \\
				
				\hline
			\multirow{1}{*}{\shortstack[l]{Compared\\ Model}}	&	ALMT & 1.080 & 0.320 \\  
			\multirow{3}{*}{MLLM}	&	GPT-4o & 0.800 & 0.600 \\
				&	Qwen2.5-vl & 2.500 & 1.100 \\
			&		VideoLLaMA3 & 2.000 & 0.600 \\
				
				\hline
		Our Model	&	\textbf{TEXT}& \textbf{1.390} & \textbf{0.010} \\
				\hline			
			\multirow{3}{*}{Uni-modal}	& \qquad			T & 1.060 & 0.340 \\
				& \qquad	A & 1.780 & 0.380 \\
				& \qquad	V & 0.870 & 0.530 \\
			\multirow{3}{*}{Two Modals}	& \qquad	T \& A & 1.350 & 0.050 \\
				& \qquad	T \& V & 1.330 & 0.070 \\
				& \qquad	A \& V & 1.510 & 0.110 \\
				
				\hline
			\multirow{3}{*}{\shortstack[l]{Uni-modal \\ w/o \\explanation}}	& \qquad			T   & 1.010 & 0.390 \\
			&\qquad	A  & -0.040 & 1.440 \\
			&\qquad	V  & -0.380 & 1.780 \\
							\hline
				\multirow{3}{*}{\shortstack[l]{Two Modals \\ w/o \\explanation}}	& \qquad	T \& A  & 1.050 & 0.350 \\
				& \qquad	T \& V   & 1.310 & 0.090 \\
 
			&	\qquad	A \& V  & 0.070 & 1.330 \\
				
				\hline
				\multirow{7}{*}{\shortstack[l]{Component\\ Ablation\tnote{1}}}&	\quad w/o explanations & 0.550 & 0.850 \\
			&	\quad  EA$\leftarrow$ Linear& 1.270 & 0.130 \\
			&	\quad  TA $\leftarrow \Circled{c}$ & 1.420 & 0.200 \\
			
			&  \quad TA $\leftarrow $   Mamba & 1.220 & 0.180 \\
			&  \quad TA $\leftarrow $   TCA & 1.080 & 0.320 \\
			
			&	\quad SMoE $\leftarrow$ Trans& 1.320 & 0.080 \\
			&	\quad w/o GF & 1.350 & 0.050 \\
				\hline
				
			\end{tabular}
		\end{threeparttable}
		\caption{Predicted scores for the case in ~\ref{fig:demo1}. The best results are highlighted in bold.   $\sigma$: deviation. See \tab\ref{table:ablationMOSEI} for more abbreviations.}
		\label{table:sample}
	\end{table}

\tab\ref{table:sample} provides evidence for three key observations. First, in this specific case, \m\ shows the strongest alignment with human judgment, with a negligible discrepancy of only 0.01. Second, the textual modality acts as the dominant information channel. For example, relying solely on textual data—even without explanatory context—still results in a relatively small discrepancy of 0.390. Third, explanations effectively compensate for the limitations of audio and visual modalities, especially for audio. When explanations are incorporated, audio-based predictions achieve a discrepancy of 0.380; by contrast, omitting explanations leads to a significant performance decline, with the discrepancy rising to 1.440. Notably, although these statistical patterns are consistent, GF and SMoE have shown minimal impact in this particular scenario. For example, without GF or SMoE, the derivation increases by 0.050.  

\subsection{Discussion}

\label{sec:discussion}

Our experiments on hyperparameters show that three layers are optimal for SMoE on \dsc. On the other hand, while MLLMs can be very good at MSA on some datasets, these datasets might be memorized by MLLMs~\cite{wang2024unlocking}. As evidence, the power of GPT-4o diminishes on Chinese datasets. Although this paper only considers MSA for Chinese/English, we will expand TEXT to use more MLLMs in multiple languages.

\section{Conclusion}

\label{sec:conclusion}

In this paper, we propose a text-routed mixture-of-experts model with explanation and temporal alignment for multi-modal sentiment analysis. With a novel temporality-oriented neural network block and cross-attention, our model performs explanation and temporal alignment. While the explanation comes from exploring the power of MLLMs, our temporal alignment block is a task-oriented neural network block design. Then, the aligned embedding is further processed by a new text-routed sparse mixture-of-experts with a gate fusion. As a result, we achieve the best performance across four datasets among all tested models, which include three state-of-the-art models and three leading MLLMs. However, as we rely on multiple MLLMs, eliminating cumulative error (e.g., from VideoLLaMA) is our future work. 

\clearpage
\bibliography{TEXT}
\clearpage

\end{document}